\documentclass{article}

\usepackage[preprint]{neurips_2024}
\usepackage[utf8]{inputenc} % allow utf-8 input
\usepackage[T1]{fontenc}    % use 8-bit T1 fonts
\usepackage{hyperref}       % hyperlinks
\usepackage{url}            % simple URL typesetting
\usepackage{booktabs}       % professional-quality tables
\usepackage{amsfonts}       % blackboard math symbols
\usepackage{nicefrac}       % compact symbols for 1/2, etc.
\usepackage{microtype}      % microtypography
\usepackage{xcolor}         % colors
\usepackage{multirow}
\usepackage{multicol}

\usepackage{graphicx}

\usepackage{amsmath}
\usepackage[most]{tcolorbox}

\usepackage{enumitem}               %%% 修改
\usepackage{makecell} 
\usepackage{subcaption}
\usepackage{ulem}
\tcbset{
    mybox/.style={
        enhanced,
        colback=teal!5!white,
        colframe=teal!80!black,
        center title,
        fonttitle=\bfseries,
        sharp corners,
        title=Instruction Template of ReAct Prompting,
    }
}

\title{Improving Generalization in Intent Detection: GRPO with Reward-Based Curriculum Sampling}

\author{%
Zihao Feng$^{1,2}$\thanks{Equal contribution}\ \  \thanks{Zihao Feng was an intern at Tencent during the preparation of this work}\ \ , Xiaoxue Wang$^{1 *}$, Ziwei Bai$^{1 *}$, Donghang Su$^{1 *}$, \\ \textbf{Bowen Wu$^{1}$, Qun Yu$^{1}$, Baoxun Wang$^{1}$} \\
  $^1$Platform and Content Group, Tencent\\
  $^2$Faculty of Computing, Harbin Institute of Technology \\
  \texttt{21b903052@stu.hit.edu.cn} \\
  \texttt{\{yukixxwang, ziweibai, ashersu, jasonbwwu, sparkyu, asulewang\}@tencent.com} \\
}

\begin{document}

\maketitle

\begin{abstract}
  Intent detection, a critical component in task-oriented dialogue (TOD) systems, faces significant challenges in adapting to the rapid influx of integrable tools with complex interrelationships. Existing approaches, such as zero-shot reformulations and LLM-based dynamic recognition, struggle with performance degradation when encountering unseen intents, leading to erroneous task routing. To enhance the model's generalization performance on unseen tasks, we employ Reinforcement Learning (RL) combined with a Reward-based Curriculum Sampling (RCS) during Group Relative Policy Optimization (GRPO) training in intent detection tasks. Experiments demonstrate that RL-trained models substantially outperform supervised fine-tuning (SFT) baselines in generalization. Besides, the introduction of the RCS, significantly bolsters the effectiveness of RL in intent detection by focusing the model on challenging cases during training. 
  Moreover, incorporating Chain-of-Thought (COT) processes in RL notably improves generalization in complex intent detection tasks, underscoring the importance of thought in challenging scenarios.
  This work advances the generalization of intent detection tasks, offering practical insights for deploying adaptable dialogue systems.
\end{abstract}

\begin{figure}[hb]
\centering
\includegraphics[width=\textwidth]{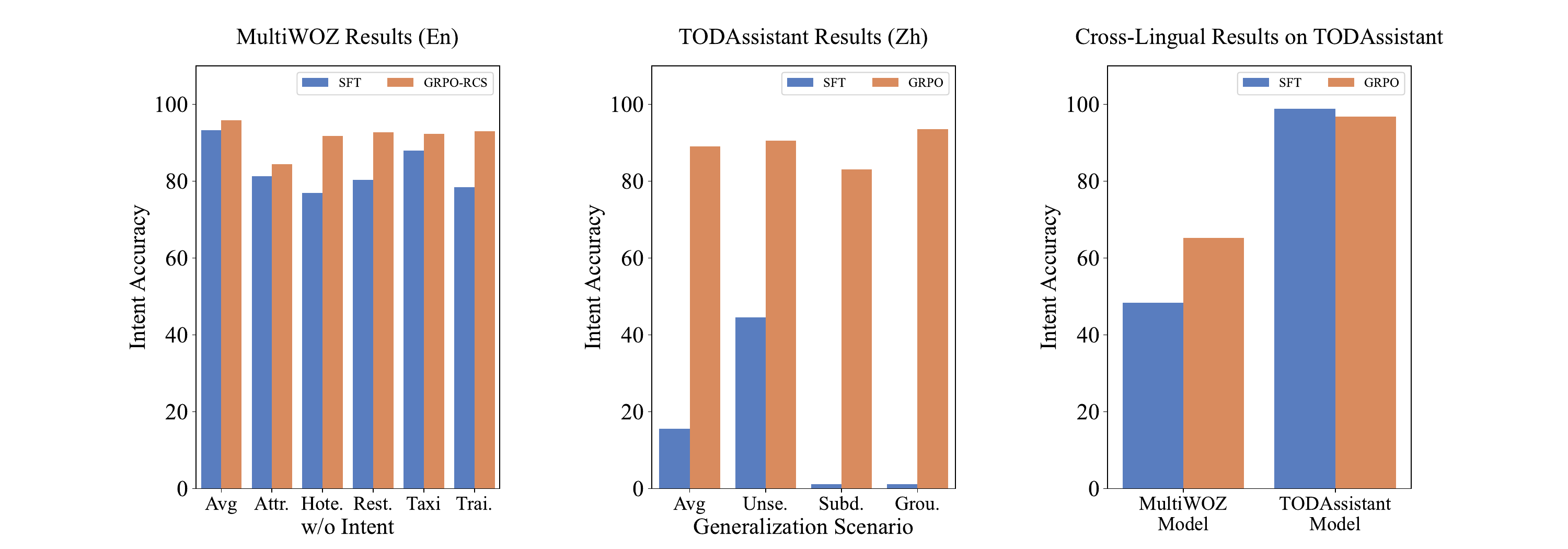}
\caption{Comparative performance of RL-Trained and SFT-Trained models in intent detection across various generalization scenarios} 
\label{firstfig}
\end{figure}

\section{Introduction}
\label{section:introduction}

%Intent detection \cite{weld2022survey, casanueva2020efficient}, a crucial component in Task-oriented dialogue (TOD) systems \cite{gupta2024dard, xu2024rethinking}, aims to identify the underlying purpose of a user's query. 
%This provides a solid foundation for subsequent components like dialogue state tracking, tool calls, dialogue management strategies and response generation.
%By enabling dynamic task allocation among agents through their interactions, intent detection ensures TOD systems can efficiently adapt to evolving task priorities and requirements.

As a crucial component of Task-oriented Dialogue (TOD) systems \cite{gupta2024dard, xu2024rethinking}, the intent detection module aims to identify the underlying requirements of users' queries \cite{weld2022survey, casanueva2020efficient}. 
Consequently, the intent detection models are expected to efficiently adapt to evolving task priorities and requirements, so as to conduct dynamic task allocation among multiple agents in complicated application scenarios. 

% With the rapid iteration of integrable API tools \cite{du2024anytool, qu2024chatgpt} in both academia and industry, the influx of new tools that can be incorporated into TOD systems is increasing. However, this growth introduces complexity in the interrelationships among these tools, with some exhibiting similar functionalities, some overlapping, and others forming subset relationships. This complexity presents new challenges in recognizing and invoking new tools, particularly as these tools have not been trained in real-time by the model. Consequently, it is imperative to enhance the generalization of intent detection model for accurate recognition and invocation of unseen tools to ensure the efficient operation of TOD systems in practical use.

The recent development of LLMs has accelerated the evolution of TOD systems, 
and with the rapid iteration of integrable artificial API tools \cite{du2024anytool, qu2024chatgpt}, the number of AI tools that can be incorporated into TOD systems is increasing. 
This situation leads to a great challenge that, actually, intent detection models need to flexibly adapt to newly introduced tools for unseen tasks, with no timely incremental training processes.
In many cases, tools within the management of intent detection modules maintain complex interrelationships, such as functional similarity, overlapping, inclusion, etc.
Thus, the generalization of intent detection models is the essence for TOD systems to adjust to complicated practical scenarios, in which a number of tools, with complex relationships and interactions, may be frequently involved.

% Efforts are underway to improve the accuracy of unseen tasks in existing methods. 
% For instance, the model proposed by Siddique et al.\cite{DBLP:conf/sigir/SiddiqueJXH21} heavily relies on external common sense knowledge. 
% In contrast, the model by Comi et al.\cite{DBLP:conf/emnlp/ComiCPM23} reformats the tasks into an NLI format to achieve zero-shot capability. 
% Moreover, LLM-based models~\cite{parikh2023exploring, gupta2024dard} dynamically recognize unknown tasks by capitalizing on their inherent zero-shot capabilities. 
% However, these models often experience significant performance degradation in intent detection models when confronted with unseen or new intent, resulting in the system incorrectly routing user intent to the wrong agent.
% Therefore, enhancing the generalization of the intent detection model is particularly critical. 

Previous studies have made much efforts to improve the accuracy of intent detection models by adopting new tools to handle unseen tasks. 
For example, the model proposed by Siddique et al. introduces external common sense knowledge to address this problem \cite{DBLP:conf/sigir/SiddiqueJXH21}. 
Comi et al.\cite{DBLP:conf/emnlp/ComiCPM23} reformatted the tasks in an NLI format to achieve zero-shot capability.
Moreover, LLM-based models~\cite{parikh2023exploring, gupta2024dard} dynamically recognized unknown tasks by capitalizing on their inherent zero-shot capabilities. 
However, these models often experienced significant performance degradation in intent detection models when confronted with unseen or new intent, resulting in the system incorrectly routing user intent to the unmatched agent.
This situation indicates that enhancing the generalization of intent detection models is particularly critical.

Reinforcement learning has been proved to be valuable in improving the generalization of LLMs \cite{swamy2025roadsleadlikelihoodvalue},
which has also been supported by the exceptional cross-task generalization of the recent model DeepSeek-R1 \cite{guo2025deepseek}. 
Inspired by the principle of DeepSeek-R1, we propose to apply the Group Relative Policy Optimization (GRPO) methodology to enhance the generalization of the intent detection model. 
In particular, to ensure that the R1-style RL process achieves expected performances on the intent detection problem, a sampling strategy is presented in this work. 
As depicted in Figure \ref{firstfig}, the experimental results demonstrate that in varying generalization scenarios, the reinforcement learning (RL) model successfully predicts user query intents, significantly outperforming the supervised fine-tuned (SFT) model. This superiority is particularly evident in terms of generalization across unseen intents, subdivided intents, grouped intents, and cross-language.
In conclusion, our work offers the following findings:

% It is well known that DeepSeek-R1 \cite{guo2025deepseek} effectively transfers the reasoning patterns learned in symbolic domains such as math, code, STEM, etc., from its base version (R1-zero) to a broader range of natural language processing tasks, demonstrating exceptional cross-task generalization ability.
% Inspired by this generalization characteristic, we propose applying the Reinforcement Learning (RL) method used in DeepSeek-R1 to the intent detection tasks, aiming to enhance its generalization. We also propose applying a Reward-based Curriculum Sampling Strategy in the Group Relative Policy Optimization (GRPO) training process for intent detection tasks. The experimental results indicate that the method significantly improves the model's generalization in the intent detection tasks. 
% In conclusion, this paper offers the following findings:

% This paper offers the following contributions:

\begin{itemize}
    \item We demonstrate that models trained with RL significantly outperform those trained with SFT on the intent detection problem,
    in terms of generalization across unseen intents, subdivided intents, grouped intents, and cross-language.
    \item To stimulate the capability of GRPO training, we introduce the Rewards-based Curriculum Sampling Strategy, 
    which is found to be valuable for enabling models to focus more on challenging cases during the training process.
    \item
    % The "thought" component plays a more crucial role during the training process in more challenging datasets.
    Incorporating COT~\cite{wei2022chain} processes during reinforcement learning significantly enhances model generalization on complex intent detection tasks, highlighting the importance of thought processes for improving generalization in challenging scenarios.
    \item Furthermore, our experiments also show that even a base model without instruction training can achieve performance comparable to the instruction model on the intent detection task. This finding suggests that the Function Call capability of the base model may not be a necessary prerequisite for intent detection models trained with RL. 
    % Furthermore, our experiments indicates that a robust Function Call capability is not a prerequisite for RL training in intent detection tasks. We demonstrate that even a base model can achieve a performance level that approximates that of the instruction model.
\end{itemize}

\section{Method}

\subsection{Task Formulation}

In task-oriented dialogue systems, accurate detection of user intents is essential for dialogue state tracking and subsequent API execution. We formulate the intent detection task as follows: Given a dialogue history \(H = \{(u_1, a_1, y_1), (u_2, a_2, y_2), \ldots, (u_{t - 1}, a_{t - 1}, y_{t - 1})\}\), where \(u_i\), \(a_i\), and \(y_i \in \mathcal{Y}\) represent the user's utterance, the assistant's response, and the ground truth intent label at turn \(i\), respectively.
\(\mathcal{Y}=\{c_1, c_2, \ldots, c_K\}\) denotes a predefined set of \(K\) actionable intents related to domain-specific operations, with each intent \(c_i\) associated with a natural language description \(d_i\) in the prompt. 
The objective of an intent detection model $M$ is to accurately predict the intent \(y_t \in \mathcal{Y}\) of the current user's utterance \(u_t\). Formulated as:
\begin{equation}
  \mathcal{L}(\theta) = -\frac{1}{N} \sum_{n=1}^{N} \log P_{\theta}(y_t^n | H^n, u_t^n, {d_1, d_2, \ldots, d_K})  
\end{equation}
where $\theta$ represents the parameters of model $M$, $N$ is the number of training examples, $P_{\theta}$ denotes the probability assigned by model $M$.

% Therefore, the model $M$ can theoretically address intent detection tasks for new intent sets through dynamic prompt engineering.
Apparently, the model $M$ demonstrates non-trivial generalization potential for evolving dialogue systems, as its architecture theoretically supports the discovery of novel intent categories through dynamic prompt engineering.
% A well-generalizing intent detection model should be able to recognize previously unseen intents by dynamically adjusting the intent descriptions in the prompt.
Formally, for $y_t = c_{K+1} \notin \mathcal{Y}$, model $M$ can add the description $d_{K+1}$ of $c_{K+1}$ to the prompt to predict $y_t$. In particular, this $y_t$ may represent not only a completely new category distinct from $\mathcal{Y}$, but also potentially a division or recombination of previous categories.

\subsection{Intent Detection via Reinforcement Learning}

% Based on the problem definition, the most direct training approach for intent detection models is to learn to predict $y_t$ given history $H$ and current utterance $u_t$ through supervised fine-tuning. 
Directly applying supervised fine-tuning (SFT) to learn the prediction of $y_t$ has been a conventional approach, however, this method often suffers from poor generalization capabilities.
In this paper, inspired by DeepSeek-R1-Zero~\cite{guo2025deepseek}, which demonstrated the significant potential of reinforcement learning combined with model reasoning, we design rule-based rewards and exclusively employ GRPO~\cite{shao2024deepseekmath} to guide model training.

%\sout{Specifically, we focus on intent recognition tasks in multi-turn task-oriented dialogues to investigate the effectiveness of this approach.}
Specifically, building upon an arbitrary LLM, we construct a complete prompt using the ReAct Prompting~\cite{yao2023react} method, where the system prompt is "You are a helpful assistant.".
In the final turn of the dialogue, we insert an instruction composed of the user query and other relevant information, such as descriptions of available tools. The specific instruction template is as follows.

\begin{tcolorbox}[mybox]
\tiny
You are an agent that helps users choose the right tool or tools from the list of given tools to solve their problems. \\
\\
For each tool, you are first given its description and required parameters. Then, a logic module specifically explains the logical information needed for this tool to handle multi-turn conversation issues.\\
\\
\#\# Tool APIs\\
\\
\texttt{\{tools text\}}\\
\\
\#\# Task Logic\\
\\
\texttt{\{logic text\}}\\
\\
\#\# Output Format\\
\\
Use the following format:\\
\\
Last Tool: the tool used in last query\\
Question: the input question you must answer\\
Thought: you should always think about what to do\\
Action: the action to take\\
Finish!\\
\\
Begin!\\
Last Tool: \texttt{\{tool\}}\\
Question: \texttt{\{query\}}
\end{tcolorbox}

% We designed a rule-based reward system to guide reinforcement learning training, which specifically includes Format Reward and Answer Reward.

Regarding the training objectives, we design two rule-based reward functions to guide reinforcement learning training. Specifically, these include a Format Reward to constrain the model's output structure and an Answer Reward to evaluate the correctness of intent detection.
\begin{equation}
R = \lambda_{\text{format}} \cdot R_{\text{format}} + \lambda_{\text{answer}} \cdot R_{\text{answer}}
\end{equation}
where $\lambda_{\text{format}}$ and $\lambda_{\text{answer}}$ are weighting coefficients for each respective reward component. 

\paragraph{Format Reward} 
We restrict the model's output to strictly follow a fixed format, as specified in the Instruction Template of ReAct Prompting. Specifically, the model's output must strictly conform to a three-line structure where each line begins with ``Thought:'', ``Action:'', and ``Finish!'' respectively. Each of these keywords must appear exactly once in the entire output and the content of the third line is limited to solely   ``Finish!''.

\[
R_{\text{format}} = 
\begin{cases}
1, & \text{if format is correct}\\
0, & \text{otherwise}
\end{cases}
\]

\paragraph{Accuracy Reward}
The accuracy-based reward is a binary metric that evaluates the exact match between the predicted intent $\hat{y}_t$ and the ground truth label $y_t$. We employ a regular expression-based method to extract the predicted intent from the model's output.

\[
R_{\text{answer}} = 
\begin{cases}
1, & \text{if the answer $\hat{y}_t$ fully matches the ground truth $y_t$}\\
0, & \text{otherwise}
\end{cases}
\]

% This mechanism guides the model's reasoning process during response generation, enabling a more precise differentiation of user intents. Compared to conventional SFT approaches, our GRPO-trained model demonstrates significantly enhanced generalization capabilities, particularly evidenced by its superior performance in optimizing intent distributions for task-oriented dialogue systems and effectively handling emerging intents through dynamic prompt engineering.

\subsection{Reward-Based Curriculum Sampling}

% We observe that the GRPO-trained model exhibits a notable convergence rate in dialog intent detection. The model converges to a comparable accuracy of the SFT model in dozens of steps. However, this rapid convergence presents a limitation: the model's focus on more challenging examples diminishes in the subsequent phases of training. To address this issue, we employ an offline Reward-based Curriculum Sampling Strategy to enhance both the efficiency and effectiveness of the training process.

Research indicates that low reward variance leads to a flat landscape in the RLHF objective, resulting in suboptimal convergence~\cite{razin2025makes}. Our observations on intent detection tasks reveal that GRPO-trained models converge remarkably quickly, reaching accuracy comparable to SFT models within dozens of training steps. Consequently, in subsequent training phases, the reward variance becomes extremely small, and the model's focus on challenging examples diminishes. To address this issue, we employ an offline Reward-based Curriculum Sampling Strategy to enhance both the efficiency and effectiveness of the training process.

\paragraph{Offline Reward Collection} 
To select the most challenging sample for RL, we first apply the GRPO method to the entire training dataset, recording the rewards for each data accross all samples throughout the GRPO training process. Just as shown in Eq~\ref{score_eq}, the $G$ represents the sampling number of each data, $R^{i,j}$ represents the reward of $j$-th sampling of the $i$-th data, and the $Score_{i}$ represents the score of $i$-th data.

% The results indicate that 79$\%$ of the data consistently achieved full scores across all samples, suggesting that a significant portion of the data becomes redundant in the later stages of training.

\begin{equation}
\label{score_eq}
    Score_{i} = \sum_{j=1}^{G} (\lambda_{\text{format}} \cdot R_{format}^{i,j} + \lambda_{\text{answer}} \cdot R_{answer}^{i,j})
\end{equation}

\paragraph{Curriculum Sampling} 
After obtaining the training rewards for each sample, we employ a two-stage training method. In the first stage, we train the model for dozens of steps on the entire dataset until the accuracy on the validation set changes less than a certain threshold. 
We intentionally avoid using easier data during this initial stage because the model demonstrated significantly low rewards across all examples at the beginning of the training process. In addition, this approach facilitates the transition of our proposed method to an online format in subsequent work. In the second stage, we define the $i$-th data is challenging when the $Score_{i} < (\lambda_{\text{format}} + \lambda_{\text{answer}})*G$.
We select the challenging data to continue training the model trained in the first stage. This approach allows the model to concentrate on these difficult examples during the second stage.

% \paragraph{Positive Instance Sampling} 
% Due to the fact that the second stage exclusively comprises erroneous data encountered during RL, the initial accuracy of the second stage is considerably low, showing a substantial deviation from the foundation model of the first stage. This deviation maybe lead to model degradation. Therefore, we randomly sampling data from the whole training set to combine with the curriculum samples in order to better maintain the performance achieved in the first stage.

% combine the incorporated positive instances with 

% \section{Experimental Setup}

% \subsection{Dataset}
% We evaluate our proposed method on the subset of MultiWOZ 2.2 dataset \cite{zang2020multiwoz}.  The dataset is a large-scale multi-domain TOD dataset which contains 10437 conversations and comprises 7 domains. 
% We extracted the intent sub-task from the dataset for training and evaluation of our proposed method.

% \subsection{Setup}
% We select Qwen2.5-7B-Instruct\footnote{\url{https://huggingface.co/Qwen/Qwen2.5-7B-Instruct}} \cite{qwen2.5}, a representative and common open source LLM as our foundation model. 
% We utilized the $10k$ data generated from MultiWOZ2.2 to conduct RL on the Qwen model.
% As for the supervised finetune method, we fully fine-tuned the model for 2 epochs.
% % We generate the training dataset tailored to each agent. All the agents are fully fine-tuned and conducted on 8 A100 GPUs with 40GB of RAM for 2 epochs. 

\section{Experimental Setup}

\subsection{Dataset}

We conduct experiments on two task-oriented dialogue datasets.

The first dataset is the widely used MultiWOZ benchmark, specifically a subset of \textbf{MultiWOZ 2.2} \cite{zang2020multiwoz}. This large-scale multi-domain task-oriented dialogue dataset contains 10,437 conversations spanning 7 domains. These domains encompass tasks that require multiple interaction turns to complete, such as flight booking and hotel reservations. We extract the intent subtask from this dataset for training and evaluation of our proposed method.

Additionally, considering the rapid iteration of integrable artificial API tools, we construct a dataset that simulates interactions with a general AI assistant that integrates various task capabilities, named \textbf{TODAssistant}. This dataset encompasses 10 task categories, including traditional task-oriented functions such as signature settings, friend recommendations, and chatbot recommendations, as well as AI-driven task types, including text-to-image generation, image style transformation, and text-based conversation. All dialogue data for these tasks were simulated using GPT-4o~\cite{hurst2024gpt} to generate conversations representing possible interaction scenarios for each tool, with specific details omitted here. In summary, this is a task-oriented dialogue dataset containing 10 tasks, covering both traditional task-oriented dialogue-focused areas and emerging AI-driven tasks. The data is entirely generated by LLMs and comprises 9,500 training samples and 500 test samples.

% To better evaluate model adaptability to situations involving new, split, or merged tasks, we further developed three out-of-domain test sets (i.e., not included in the known 10 categories):
To better evaluate model adaptability to situations involving new domains, subdivided, or grouped tasks, we further develop three generalization test sets with new intents that are not included in the known 10 categories:

\begin{itemize}[leftmargin=0.5cm]

\item \textbf{TODAssistant-Unseen5}: Introduces 5 completely novel tasks not encountered in the dataset, including singing children's songs and storytelling, which are oriented toward children's scenarios.

\item \textbf{TODAssistant-Subdivided}: For the text chat task already included in the 10 categories, we divide it into three more granular intents to simulate real-world scenarios where finer-grained capabilities might better address specific user needs. Specifically, we split the text chat task into:

\begin{itemize}
    \item Various text processing intents: Covering purpose-specific text generation tasks including translation, text classification, text generation, mathematical calculation, and code generation.
    \item Safety topics: Involving content related to pornography, violence, etc. 
    \item Free topic conversation: Chit-chat or intents not belonging to the other two categories.
\end{itemize}

\item \textbf{TODAssistant-Grouped}: This set simulates situations where, due to agent upgrades, multiple previously separate tasks may be completed by a single agent. Specifically, we regroup two relatively similar intents — "friend recommendations" and "chatbot recommendations" into a single intent.

\end{itemize}

To clarify, TODAssistant-Unseen5 introduces 5 entirely new task categories, TODAssistant-Subdivided uses a portion of the test samples originally belonging to the text chat task and divides them into three new intents, and TODAssistant-Grouped modifies the intent of two test set categories into one new intent. It is important to emphasize that none of these categories were encountered during the training process.

\subsection{Setup}

We selected Qwen2.5-7B-Instruct\footnote{\url{https://huggingface.co/Qwen/Qwen2.5-7B-Instruct}} \cite{yang2024qwen2} as our foundation model, which represents a widely adopted open-source large language model. 

For the MultiWOZ2.2 dataset, we utilize the $10k$ conversations to conduct reinforcement learning. We conduct 60 steps for the first stage of the curriculum learning, and 1 epoch (153 steps) for the second stage. For both of the two stages, we train our model with a learning rage of $9.0*10^{-6}$, incorporating a sampling strategy that generated 7 responses per prompt at a temperature parameter of 0.7.
In the case of the TODAssistant dataset, we employ the entire training set for our experiments. We train the model with a learning rage of $3.0*10^{-6}$, incorporating a sampling strategy that generated 7 responses per prompt at a temperature parameter of 0.9.
For all the datasets, we utilize a global batch size of 448 for our training.

Regarding the supervised fine-tuning approach, we fully fine-tune the model with the same epoch of the corresponding GRPO-based method. On both datasets, we employ \textbf{Accuracy} as the metric to measure the effectiveness of intent detection.

\section{Experiments}

\subsection{Comparison of Reinforcement Learning and Supervised Fine-Tuning Effects}

We conduct intent detection training on two datasets using both GRPO and SFT approaches. Our evaluation strategy involves testing in-domain intent categories (those present in the training data) and out-of-domain intent categories (those unseen during training). It is important to note that the GRPO training discussed in this subsection corresponds to the methodology described in Section 2.1, which does not incorporate curriculum learning. Our primary objective is to analyze the performance differences between models trained using GRPO versus those trained through standard SFT.

\subsubsection{Performance on In-Domain Test Set}

As shown in Table~\ref{main_results_simple_task}, both SFT and GRPO-trained models significantly improve intent recognition performance on in-domain categories. However, using only RL (GRPO) on the same training data as SFT does not surpass SFT's performance on in-domain testing. While both approaches achieve comparable convergence results on the more complex MultiWOZ 2.2 dataset, GRPO performs slightly worse on the machine-generated TODAssistant dataset.

\begin{table}[t]
  \caption{Results of the in-domain evaluation on two datasets.}
  \label{main_results_simple_task}
  \centering
  \begin{tabular}{lcccccc}
    \toprule
    Model &  TODAssistant & MultiWOZ 2.2 & Avg   \\
    \midrule
    Qwen2.5-7B-Instruct  &  22.4  & 23.2 & 22.8  \\
    Qwen2.5-7B-Instruct + SFT & 98.8  &  93.3 & 96.1 \\
    Qwen2.5-7B-Instruct + GRPO &  96.8 & 93.3 & 95.1 \\ 
    \bottomrule
  \end{tabular}
\end{table}

% \subsubsection{Performance on Out-of-Domain Test Set}

\begin{table}[t]
  \caption{Results of the out-of-domain evaluation on MultiWOZ 2.2 dataset. The symbol ``$^\dagger$'' denotes the performance on the excluded intent category that was unseen in the training data.}
  \label{main}
  \centering
  \resizebox{\textwidth}{!}{
  \begin{tabular}{lcccccc}
    \toprule
    Model & Attraction & Hotel & Restaurant & Taxi & Train & Avg \\
    \midrule
    \multicolumn{7}{l}{\emph{\underline{Baseline}}}    \\
    Qwen2.5-7B-Instruct + SFT &  92.6  &  93.5  & 93.4 & 95.6 &  92.3 & 93.3 \\
    Qwen2.5-7B-Instruct + GRPO     & 94.5  &  91.6 & 91.9 & 93.9 & 92.6  &  93.3  \\ \midrule
    \multicolumn{7}{l}{\emph{\underline{w/o Attraction}}}    \\
    Qwen2.5-7B-Instruct + SFT &  \textbf{\textit{43.8}}$^\dagger$ &  \underline{94.3}  & \underline{93.7} & 96.4 & 92.9  & 81.3 \\
    Qwen2.5-7B-Instruct + GRPO     & 43.1$^\dagger$  &  92.7 & 93.0 & \underline{97.5} & \underline{93.3}  &  \underline{84.4}  \\ \midrule
    \multicolumn{7}{l}{\emph{\underline{w/o Hotel}}}    \\
    Qwen2.5-7B-Instruct + SFT &  93.5 &  37.1$^\dagger$  & \underline{92.3} & 95.0 & 91.3  & 76.9 \\
    Qwen2.5-7B-Instruct + GRPO     & \underline{95.3}  &  \textbf{\textit{87.1}}$^\dagger$ & 92.3 & \underline{96.1} & \underline{92.6}  & \underline{91.8}  \\ \midrule
    \multicolumn{7}{l}{\emph{\underline{w/o Restaurant}}}    \\
    Qwen2.5-7B-Instruct + SFT &  92.6 & 89.7   & 57.1$^\dagger$ & 93.6 &  \underline{92.1} & 80.3 \\
    Qwen2.5-7B-Instruct + GRPO     & \underline{95.1}  &  \underline{93.0} & \textbf{\textit{91.2}}$^\dagger$ & \underline{95.3} & 91.9  &  \underline{92.8}  \\ \midrule
    \multicolumn{7}{l}{\emph{\underline{w/o Taxi}}}    \\
    Qwen2.5-7B-Instruct + SFT & 87.0  & 90.0   & 92.7 & 53.4$^\dagger$   & 89.6 & 88.0 \\
    Qwen2.5-7B-Instruct + GRPO     & \underline{95.9}  &  \underline{92.5} & \underline{92.6} & \textbf{\textit{74.2}}$^\dagger$ & \underline{92.9}  &  \underline{92.3}  \\ \midrule
    \multicolumn{7}{l}{\emph{\underline{w/o Train}}}    \\
    Qwen2.5-7B-Instruct + SFT & 92.1  &  91.1  & \underline{94.1} & 91.8 & 47.9$^\dagger$  & 78.4 \\
    Qwen2.5-7B-Instruct + GRPO     & \underline{95.9}  &  \underline{93.1} & 92.6 & \underline{96.8} & \textbf{\textit{90.6}}$^\dagger$  &  \underline{93.0}  \\ 
    \bottomrule
  \end{tabular}}
\end{table}

% \begin{table}[htbp]
%   \caption{Results of the out-of-domain evaluation on TODAssistant dataset}
%   \label{main_results_simple_task_out}
%   \centering
%   \resizebox{\textwidth}{!}{
%   \begin{tabular}{lccccc}
%     \toprule
%     Model & TODAssistant & Unseen5 & Subdivided & Grouped & Avg   \\
%     \midrule
%     Qwen2.5-7B-Instruct  & - &  63.0  & 40.2 & 21.6 & 41.6 \\
%     Qwen2.5-7B-Instruct + SFT & - & 44.5 & 0.0 & 0.0 & 14.8 \\
%     Qwen2.5-7B-Instruct + GRPO & - &  \textbf{90.6}  & \textbf{83.1} & \textbf{93.6} & \textbf{89.1} \\ 
%     Qwen2.5-7B-Instruct + GRPO (MultiWOZ) & 65.2 & - & - & - & - \\
%     \bottomrule
%   \end{tabular}}
% \end{table}

\subsubsection{Performance in generalization scenarios}
% To further analyze the advantages of RL, we evaluate the performance of models trained with SFT and RL on the unseen domain. The experiment on the TODAssistant dataset in Table \ref{main_results_simple_task_out} shows that, the SFT-trained model resulted in a sharp performance decline on out-of-domain test sets, with classification accuracy collapsing to 0\% due to severe overfitting. In contrast, the GRPO-trained model demonstrated markedly enhanced generalization capabilities on out-of-domain test sets, achieving particularly pronounced improvement.

% However, a substantial performance gap emerged between the two methods when tested on out-of-domain scenarios. 
To assess the performance of RL methodologies across various generalization scenarios, we conduct a comparative analysis of the GRPO model and the SFT model, focusing on their adaptability as the intent label set progressively evolves and deviates from the training dataset.
% Since we constructed three out-of-domain test sets for the TODAssistant dataset, we first compared the models' performance on this dataset. 
% We first compared the performance on TODAssistant dataset that includes three test set to evaluate the generalization ability in different scenarios.
% As shown in Table~\ref{main_results_simple_task_out}, SFT-trained models performed worse on all three test sets compared to their pre-training performance. 

Table~\ref{main_results_simple_task_out} shows performance on the three generalization test sets of TODAssistant.
% While the untuned Qwen2.5-7B-Instruct model demonstrated reasonable comprehension capabilities on several tasks, after training, its ability to understand unseen intents in the Unseen5 test set deteriorated. 
% For the Subdivided and Grouped test sets, which involved dividing or regrouping the original 10 training classes, the SFT-trained model continued to predict only from the 10 seen categories despite changes in input prompts, disregarding the provided instructions. 
% This resulted in the SFT model's complete failure to make correct predictions on these two test sets.
Compared to the untuned Qwen2.5-7B-Instruct model, the performance of the SFT model shows a notable decline across all three test sets.
This deterioration is especially evident on the Subdivided and Grouped test sets, where the SFT-trained model limits its predictions to the 10 categories seen during training, rather than producing new labels as instructed by the input prompts.
It suggested that the SFT model primarily learned a straightforward mapping from user queries to intent labels.
In contrast, models trained with GRPO demonstrate significant improvements across all three test sets, maintaining over 90\% accuracy on both the Unseen5 and Grouped tests.
These results indicate that the GRPO model effectively learns instruction understanding and reasoning, leading to superior generalization capabilities.
In order to further validate the above findings, we conduct additional generalization testing on the MultiWoz 2.2 dataset. 
% Specifically, when training with either SFT or GRPO, we excluded all data containing a specific intent and then evaluated model performance on both this unseen class and other classes during testing. Results are presented in Table~\ref{main}.
Specifically, we entirely exclude the data corresponding to a particular intent from the training set and then evaluate the model on the official test set, which includes both the unseen category and other categories. 
As illustrated in Table~\ref{main}, models trained with GRPO surpass those trained with SFT by over 20\% in most categories, except on the "Attraction" category where both methods yield subpar performance.
These findings underscore that GRPO training improves the generalization capability for intent detection tasks.

% Examining the diagonal results from the second through sixth columns, which represent performance on the excluded unseen classes, we observe that except for the Attraction category where both methods achieved low accuracy, GRPO-trained models outperformed SFT-trained models by more than 20\% on all the other categories.This observation further confirms GRPO training's enhancement of model generalization capability for intent detection tasks.

Interestingly, when excluding an intent categories, models trained with GRPO demonstrated stronger in-domain performance than those fine-tuned through SFT - a finding that contrasts with the primary results shown in Table~\ref{main_results_simple_task}. 
This divergence suggests that SFT models exhibit greater sensitivity to reductions in training data diversity and sample size, while GRPO-trained models maintain more consistent robustness.
Specifically, category removal leads to performance declines of 5\%-17\% in SFT models, whereas GRPO models maintain stable performance, with accuracy reductions remaining consistently below 2\% in most cases. 

% Comparing both tables reveals that SFT performance consistently declined substantially when data scale and diversity decreased due to category removal. The smallest impact occurred when removing the Taxi category, yet still resulted in a performance drop of over 5\%, while the largest decline approached 17\%. In contrast, except for the significant impact observed when excluding the Attraction category, removing other categories had minimal effect on the GRPO model's overall performance (accuracy decreased by less than 2\%).

\begin{table}[t]
  \caption{Results of the out-of-domain evaluation on TODAssistant dataset}
  \label{main_results_simple_task_out}
  \centering
  \resizebox{\textwidth}{!}{
  \begin{tabular}{lccccc}
    \toprule
    Model & TODAssistant & Unseen5 & Subdivided & Grouped & Avg   \\
    \midrule
    Qwen2.5-7B-Instruct  & - &  63.0  & 40.2 & 21.6 & 41.6 \\
    \quad + SFT & - & 44.5 & 0.0 & 0.0 & 14.8 \\
    \quad + GRPO & - &  \textbf{90.6}  & \textbf{83.1} & \textbf{93.6} & \textbf{89.1} \\ 
    \quad + GRPO (MultiWOZ) & 65.2 & - & - & - & - \\
    \bottomrule
  \end{tabular}}
\end{table}

To further validate the generalization capabilities of the GRPO method, we design and execute a rigorous cross-domain experiment, as summarized in Table \ref{main_results_simple_task_out}. 
Specifically, we train a model exclusively on the MultiWOZ dataset and subsequently evaluate its zero-shot performance on the TODAssistant corpus. 
Notably, TODAssistant presents a distinct challenge as an artificially generated Chinese dialogue dataset, differing fundamentally from MultiWOZ in both linguistic structure (Chinese vs. English) and data provenance (machine-generated vs. human-curated).
The results demonstrate that the GRPO approach maintains robustness even in such challenging cross-lingual and cross-task scenarios, thereby highlighting its superiority over models trained by SFT method.

% It's worth noting that TODAssistant is a synthesized dataset in Chinese, while MultiWOZ is a natural dialogue dataset in English. The results indicated that the GRPO method retains the proficiency of the model even amidst this challenging cross-lingual, cross-task scenario, thereby highlighting its superiority over the model trained via the SFT method. 
% In conclusion, comparative experiments across two datasets using two different testing methodologies demonstrate that GRPO training similar to R1 maintains superior generalization capabilities. While SFT training performs well on in-domain tests, it exhibits poor adaptability to adjustments in the intent set during task-oriented dialogue applications.

% In conclusion, we conduct comparative experiments across diverse test sets demonstrate that GRPO approach(similar to R1)  maintains superior generalization capabilities，while SFT training performs well on in-domain tests, it exhibits poor adaptability to adjustments in the intent set during task-oriented dialogue applications. 
In conclusion, our comprehensive comparative analysis across diverse test sets demonstrates that the GRPO approach (similar to R1) consistently maintains robust generalization capabilities. 
While SFT achieves competitive performance on in-domain evaluations, this method exhibits significant limitations in practical task-oriented dialogue scenarios, particularly when faced with dynamic adjustments to the intent schema or novel domain adaptations.

\subsection{Results of Reward-based Curriculum Sampling}

\subsubsection{Results of Curriculum Method}

% To better understand the impact of our proposed Reward-based Curriculum Sampling (RCS) method, we compare it with the SFT method and the original GRPO method, presenting the results in Table \ref{Curriculum_main}. In the first stage of our RCS method, we conduct 60 steps (compared to 150 steps for the original GRPO method) and achieve performance comparable to that of the GRPO method. Moreover, our proposed method enables the original GRPO to surpass the SFT method in the second stage. Furthermore, throughout all the training stages of the RCS method, we only utilize $80\%$ of the entire training set compared to the SFT and GRPO methods, yet achieve the best performance. The results demonstrate that, in GRPO training, the easier data are redundant, which makes it difficult for the model to focus on error-prone and more challenging cases. Our proposed method effectively addresses this issue.

To better understand the impact of our proposed Reward-based Curriculum Sampling (RCS) method, we conduct a comparative analysis against both the SFT method and the original GRPO approach, with results presented in Table~\ref{Curriculum_main}.
The first stage of our RCS method requires only 60 training steps—significantly fewer than the 150 steps needed for the original GRPO method—yet achieves comparable performance outcomes.
We therefore deliberately terminate the first stage of training at 60 steps to transition to the subsequent curriculum-based phase.
Notably, our proposed methodology enables the original GRPO to exceed SFT performance during the second training stage.
What is particularly significant is that throughout all training phases, RCS utilizes merely 60\% of the complete training dataset compared to the full dataset employed by both SFT and GRPO methods, while still delivering superior performance.
These findings suggest that easier examples within the GRPO training framework introduce redundancy, potentially hindering the model's ability to concentrate on error-prone and more challenging cases.
Our RCS method effectively addresses this limitation by prioritizing more informative training instances.

% \begin{table}[htbp]
%   \caption{Results of our proposed RCS method on MultiWOZ dataset.}
%   \label{Curriculum_main}
%   \centering
%   \resizebox{\textwidth}{!}{
%   \begin{tabular}{lcccccc}
%     \toprule
%     Model & Attraction & Hotel & Restaurant & Taxi & Train  & Avg \\
%     \midrule
%     Qwen2.5-7B-Instruct + SFT & 92.64  &  93.45  & 93.41 & 95.62 &  92.30 & 93.32 \\
%     Qwen2.5-7B-Instruct + GRPO     & 94.46  &  91.55 & 91.94 & 93.91 & 92.55  &  93.25  \\
%     Qwen2.5-7B-Instruct + GRPO + RCS (First Stage)    &  94.57   & 91.85 & 92.33 & \textbf{96.06} & 91.70 &  92.62  \\
%     Qwen2.5-7B-Instruct + GRPO + RCS (Second Stage)    &  \textbf{96.24}   & \textbf{94.79} & \textbf{94.65} & 95.70 & \textbf{94.58} & \textbf{95.89}  \\
%     \bottomrule
%   \end{tabular}}
% \end{table}

% \begin{table}[htbp]
%   \caption{Results of our proposed RCS method on MultiWOZ dataset.}
%   \label{Curriculum_main}
%   \centering
%   \resizebox{\textwidth}{!}{
%   \begin{tabular}{lcccccc}
%     \toprule
%     Model & Attraction & Hotel & Restaurant & Taxi & Train  & Avg \\
%     \midrule
%     Qwen2.5-7B-Instruct + SFT & 92.6  &  93.5  & 93.4 & 95.6 &  92.3 & 93.3 \\
%     Qwen2.5-7B-Instruct + GRPO     & 94.5  &  91.6 & 91.9 & 93.9 & 92.6  &  93.3  \\
%     Qwen2.5-7B-Instruct + GRPO + RCS (First Stage)    &  94.6   & 91.9 & 92.3 & \textbf{96.1} & 91.7 &  92.6  \\
%     Qwen2.5-7B-Instruct + GRPO + RCS (Second Stage)    &  \textbf{96.2}   & \textbf{94.8} & \textbf{94.7} & 95.7 & \textbf{94.6} & \textbf{96.0}  \\
%     \bottomrule
%   \end{tabular}}
% \end{table}

\begin{table}[t]
  \caption{Results of our proposed RCS method on the MultiWOZ dataset.}
  \label{Curriculum_main}
  \centering
  \resizebox{\textwidth}{!}{
  \begin{tabular}{lcccccc}
    \toprule
    Model & Attraction & Hotel & Restaurant & Taxi & Train  & Avg \\
    \midrule
    Qwen2.5-7B-Instruct \\
    \quad + SFT & 92.6  &  93.5  & 93.4 & 95.6 &  92.3 & 93.3 \\
    \quad + GRPO     & 94.5  &  91.6 & 91.9 & 93.9 & 92.6  &  93.3  \\
    \quad + GRPO + RCS (First Stage)    &  94.6   & 91.9 & 92.3 & \textbf{96.1} & 91.7 &  92.6  \\
    \quad + GRPO + RCS (Second Stage)    &  \textbf{96.2}   & \textbf{94.8} & \textbf{94.7} & 95.7 & \textbf{94.6} & \textbf{96.0}  \\
    \bottomrule
  \end{tabular}}
\end{table}

\begin{figure}[t]
\centering
\includegraphics[width=\textwidth]{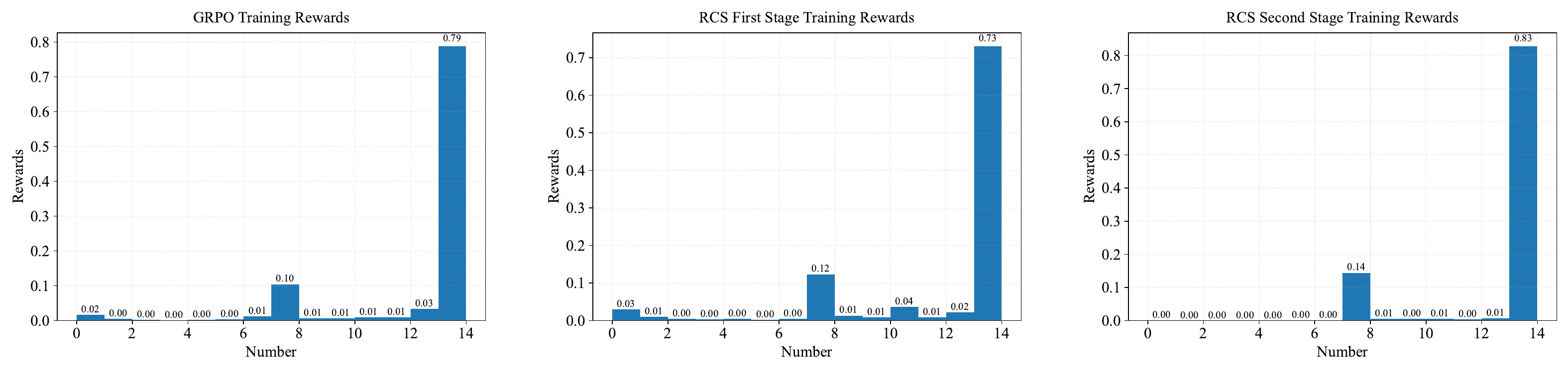} 
\caption{Histogram of rewards during the training process.} 
\label{rewardfig}
\end{figure}

To facilitate a clearer analysis of the RCS method, we present the distribution of rewards across all training data for different methods throughout the training process in Figure \ref{rewardfig}.
For each data point, we design two reward metrics and sampled seven instances, yielding a maximum possible score of 14 points per data point in the graph.
The results reveal that, compared to the original GRPO method, the RCS-based GRPO training strategy increases the proportion of perfect-score examples during the second stage, even when processing more challenging data.
These experimental findings demonstrate that the \textbf{Reward-based Curriculum Sampling Strategy enables the model to focus more effectively on challenging examples}, thereby enhancing overall model performance.

\subsubsection{Result of Positive Instance Sampling}

% To preserve the model's performance from the first stage of training, we introduce a method of positive instance sampling. The impact of different ratios of these two sampling methods is compared in Table \ref{Curriculum_pis}.

% First, we evaluate the performance of training solely with challenging examples. Besides, we combine positive data with challenging data to preserve the original performance of the model. Experimental results indicate that \textbf{training only with the challenging data does not induce degradation issues}. Furthermore, the results indicate that as the number of positive examples increases, there is a decline in model performance. The results demonstrates that \textbf{the proportion of challenging data in the second phase significantly influences the model's focus on this data}.

Due to the fact that the second stage of Curriculum Learning exclusively comprises erroneous data encountered, the initial accuracy of the second stage is considerably low, showing a substantial deviation from the foundation model of the first stage. This deviation may lead to a risk of model degradation. 
Therefore, we randomly sample data from the whole training set as \textbf{positive instance} to combine with the curriculum samples in order to evaluate the performance.

Table~\ref{Curriculum_pis} presents a comprehensive comparison of different ratios between positive and challenging samples in our curriculum.
% Our experimental design first evaluated the effectiveness of training exclusively with challenging examples, addressing concerns about potential catastrophic forgetting. 
We first evaluate the effectiveness of training exclusively with challenging examples, addressing concerns about potential catastrophic forgetting. 
Contrary to conventional beliefs, our results reveal that \textbf{exclusive training with challenging data does not lead to the expected performance degradation issues}. Furthermore, we systematically vary the proportion of positive examples in the training mixture, observing a clear inverse relationship between the percentage of positive examples and the overall performance of the model. This finding strongly suggests that \textbf{the relative concentration of challenging data in the second training phase critically determines the model's capacity to address difficult cases}. The curriculum's effectiveness appears to depend not only on the inclusion of challenging samples but on ensuring that they constitute a substantial proportion of the training distribution, allowing sufficient learning signal for the model to improve on precisely those examples where performance gains are most valuable.

\begin{table}[htbp]
  \caption{Results on different ratios between challenging data and positive data during the sampling process.}
  \label{Curriculum_pis}
  \centering
  \resizebox{\textwidth}{!}{
  \begin{tabular}{lcccccc}
    \toprule
    Model & Attraction & Hotel & Restaurant & Taxi & Train & Avg \\
    \midrule
    Qwen2.5-7B-Instruct + GRPO + RCS (1:2) & 97.0 & 94.6 & 94.0 & 96.1 & 94.1 & 94.8 \\
    Qwen2.5-7B-Instruct + GRPO + RCS (1:1) & 96.2 & 94.8 & 94.7 & 95.7 & \textbf{94.6} & 95.0  \\
    Qwen2.5-7B-Instruct + GRPO + RCS (2:1) & 96.7 & \textbf{95.3} & 95.0 & 96.8 & \textbf{94.6} & 95.4  \\
    Qwen2.5-7B-Instruct + GRPO + RCS (1:0) & \textbf{98.2} & 94.9  & \textbf{96.4} & \textbf{98.6} & 94.4 &  \textbf{96.0} \\
    \bottomrule
  \end{tabular}}
\end{table}

\subsection{Evaluating the Effect of "Thought"}

% To assess the impact of the "think" component during training on a relatively simpler task, such as intent detection, as compared to other tasks like math, we eliminated the "think" component and compared it with the model trained using the original method. We evaluate on both of these two datasets. 
% The results in Table \ref{think_table_simple_task} show that, the decline of the model's performance is not pronounced on simpler dataset. As for the more challenging dataset, the removal of the "think" component resulted in a significant drop in model performance by 17.17$\%$, underscoring that \textbf{the "think" component plays a more crucial role during the training process in more challenging datasets}.

Considering that intent detection is inherently simpler than tasks like math or coding, we investigate whether incorporating thought processes during reinforcement learning (which we term "Thought"), similar to the R1 training methodology, is truly necessary. To explore this question, we remove the "Thought"-related format loss and instructions from our reinforcement learning process and observe the resulting performance changes. We conduct experiments on both datasets.

The results in Table~\ref{think_table_simple_task} demonstrate that on the TODAssistant dataset, models without thought processes performed better on in-distribution tests, with results more closely matching those achieved after SFT. However, these models exhibit significantly reduced generalizability. However, compared to pre-trained models and SFT-trained models, their generalization ability still shows substantial improvement, \textbf{indicating that the reinforcement learning methodology itself provides inherent benefits to model generalization beyond what SFT can achieve}.

% For the MultiWOZ dataset, we observe markedly different results, with performance on benchmark tests declining considerably when thought processes were removed. 
For the MultiWOZ dataset, we observe markedly different results that the performance declining considerably as thought processes are removed.
We attribute this difference to the inherent characteristics of the two datasets: TODAssistant contains machine-synthesized data, resulting in statistically similar distributions between the training and testing sets. In contrast, MultiWOZ is a human-constructed dataset specifically designed to evaluate task-oriented dialogue capabilities, demanding a stronger understanding of known intents and better generalization to varied expressions.

Our analysis of model output lengths provides additional evidence for this disparity of difficulty: models trained on TODAssistant data generate responses averaging 37 tokens in length, while MultiWOZ-trained models produce significantly longer outputs, averaging 56 tokens. This quantitative difference further confirms the variation in task complexity between the datasets.
Consequently, \textbf{the thought process appears more beneficial for MultiWOZ (i.e., more challenging intent detection tasks) as it helps models learn recognition logic under reinforcement learning guidance}.

% \begin{table}[t]
%   \caption{Ablation results on the "Thought" during the GRPO training process.}
%   \label{think_table_simple_task}
%   \centering
%   \resizebox{\textwidth}{!}{
%   \begin{tabular}{lcccccc}
%     \toprule
%     \multirow{2}{*}{Model} & \multicolumn{4}{c}{TODAssistant} & \multirow{2}{*}{MultiWOZ2.2} & \multirow{2}{*}{Avg} \\
%     \cmidrule(lr){2-5}
%            & in-domain & Unseen5 & Subdivided & Grouped &  &  \\
%     \midrule
%     \makecell{\\ Qwen2.5-7B-Instruct + GRPO \\ (w/o think)}    &\textbf{97.8} &  86.4  & 72.7 & \textbf{94.4} & 76.1 & 85.5 \\
%     Qwen2.5-7B-Instruct + GRPO     & 96.8 &  \textbf{90.6 } & \textbf{83.1} & 93.6  & \textbf{93.3} &  \textbf{91.5}\\
%     \bottomrule
%   \end{tabular}}
% \end{table}

\begin{table}[t]
  \caption{Ablation results on the "Thought" during the GRPO training process.}
  \label{think_table_simple_task}
  \centering
  \resizebox{\textwidth}{!}{
  \begin{tabular}{lcccccc}
    \toprule
    \multirow{2}{*}{Model} & \multicolumn{4}{c}{TODAssistant} & \multirow{2}{*}{MultiWOZ2.2} & \multirow{2}{*}{Avg} \\
    \cmidrule(lr){2-5}
           & in-domain & Unseen5 & Subdivided & Grouped &  &  \\
    \midrule
    Qwen2.5-7B-Instruct + GRPO   \\
    \quad\quad  w/o think &\textbf{97.8} &  86.4  & 72.7 & \textbf{94.4} & 76.1 & 85.5 \\
    \quad\quad w/ think    & 96.8 &  \textbf{90.6 } & \textbf{83.1} & 93.6  & \textbf{93.3} &  \textbf{91.5}\\
    \bottomrule
  \end{tabular}}
\end{table}

\subsection{Base Model or Instruction Model}
% To compare the performance between the base model and the instruct model, we train models under identical experimental settings using the original GRPO method. As shown in Table~\ref{base_table}, \textbf{the base model achieved a performance comparable to the instruct model on the intent detection task}, which contradicts our prior expectations. We also present the rewards and completion lengths during the training process for both models in Figure~\ref{fig:base} and~\ref{fig:instruct}, respectively. The results suggest that while the base model converges more slowly than the instruct model, it exhibits longer completion lengths.

\begin{figure}[t]
\centering
\begin{subfigure}[t]{0.48\textwidth}
\centering
\includegraphics[width=6.5cm]{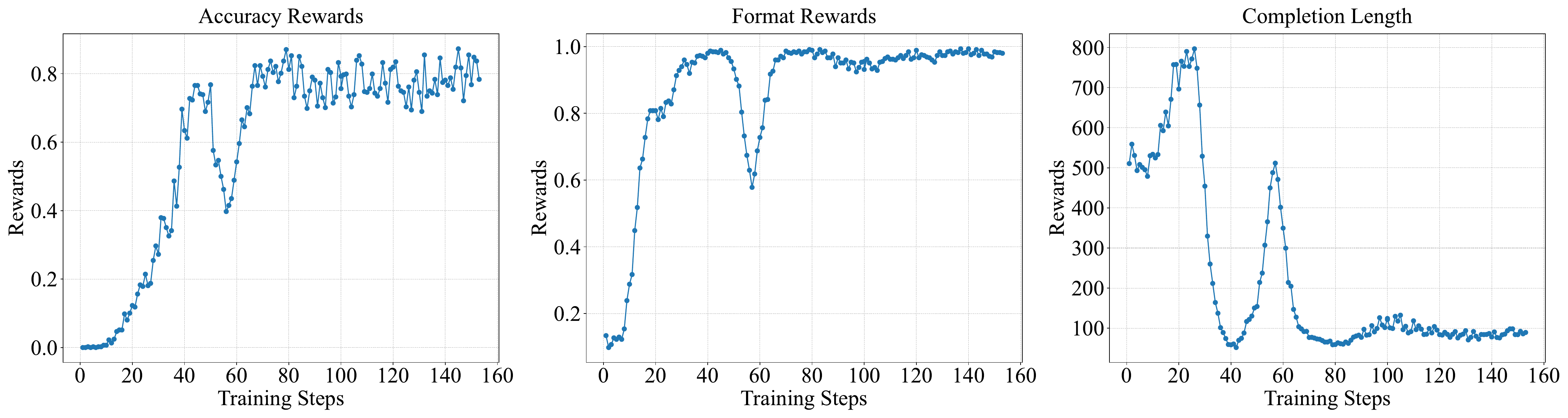}
\caption{Training states of the base model.}
\label{fig:base}
\end{subfigure}
\hfill % 用于在两个子图之间添加间距
\begin{subfigure}[t]{0.48\textwidth}
\centering
\includegraphics[width=6.5cm]{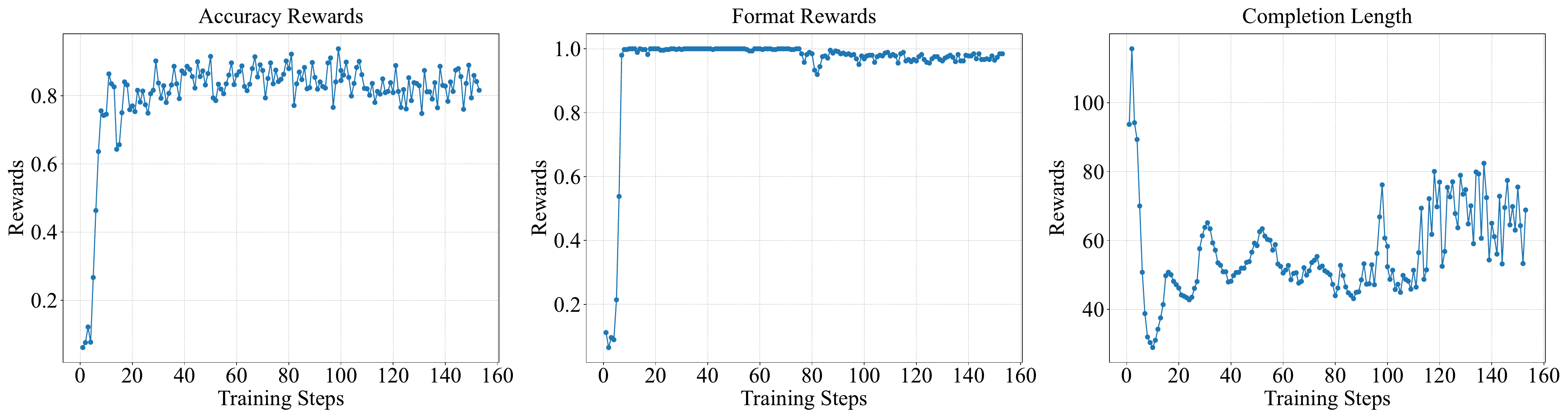}
\caption{Training states of the instruct model.}
\label{fig:instruct}
\end{subfigure}
\caption{Training curves including the accuracy, format reward, and completion length for various models.}
\label{fig:combined_training_states}
\end{figure}

Since intent detection requires models to have strong task comprehension and classification capabilities, it shares many similarities with function call tasks. Given that instruct models undergo extensive alignment training to better understand and differentiate tools, we are curious whether these models, which demonstrate significant performance improvements on function call tasks compared to base models, will also show superior results on intent detection tasks after RL training. Surprisingly, our findings align with observations from mathematical tasks: \textbf{the base model achieved performance comparable to the instruct model on the intent detection task}, as shown in Table~\ref{base_table}.
We present a comparison of rewards and completion lengths during the training process for both models in Figure~\ref{fig:base} and~\ref{fig:instruct}. Notably, while the base model converges more slowly, it ultimately achieves comparably strong performance. This discovery seems to confirm that model capabilities are primarily acquired during pre-training, with subsequent training merely helping models better utilize their inherent abilities.

\begin{table}[t]
  \caption{Results of the base model and the instruct model trained with GRPO on the MultiWOZ dataset.}
  \label{base_table}
  \centering
  \resizebox{\textwidth}{!}{
  \begin{tabular}{lcccccc}
    \toprule
    Model & Attraction & Hotel & Restaurant & Taxi & Train & Avg \\
    \midrule
    Qwen2.5-7B + GRPO  & \textbf{94.98} &  88.98 & \textbf{92.09} & \textbf{93.91} & 92.09 & 91.93 \\
    Qwen2.5-7B-Instruct + GRPO     & 94.46  &  \textbf{91.5}5 & 91.94 & \textbf{93.91} & \textbf{92.55}  &  \textbf{93.25}  \\
    \bottomrule
  \end{tabular}}
\end{table}

% To further investigate the completion lengths of the models, we reduce the learning rate and increase the training epochs for both two models. 
% We implemented two types of format rewards: 1) A strict format that rigidly restricts the output to the prescribed content, prohibiting any superfluous information; 2) A relaxed format, where the output is deemed correct as long as it encompasses the specified content. As shown in Figure~\ref{fig:base_compare} and~\ref{fig:instruct_compare}, the completion length of the instruct model remained constant under both reward functions. However, the base model displayed an initial decrease followed by an increase in completion length under the relaxed format reward, mirroring the "aha moment" described in~\cite{guo2025deepseek}. This phenomenon was absent under the stricter format reward. Despite the additional output primarily comprising task-irrelevant content, the following conclusions can still be drawn: 1) \textbf{The base model is more susceptible to experiencing "aha moments" during training compared to the instruct model}. 2) \textbf{The intent detection task may be less prone to exhibit "aha moments" due to its relative simplicity}.

% \begin{figure}[htbp]
% \centering
% \begin{minipage}[t]{0.48\textwidth}
% \centering
% \includegraphics[width=6cm]{figure/base_compare.pdf}
% \caption{Completion lengths of the base model with different format rewards.}
% \label{base_compare}
% \end{minipage}
% \begin{minipage}[t]{0.48\textwidth}
% \centering
% \includegraphics[width=6cm]{figure/instruct_compare.pdf}
% \caption{Completion lengths of the instruct model with different format rewards.}
% \label{instruct_compare}
% \end{minipage}
% \end{figure}

\begin{figure}[htbp]
\centering
\begin{subfigure}[t]{0.48\textwidth}
\centering
\includegraphics[width=6cm]{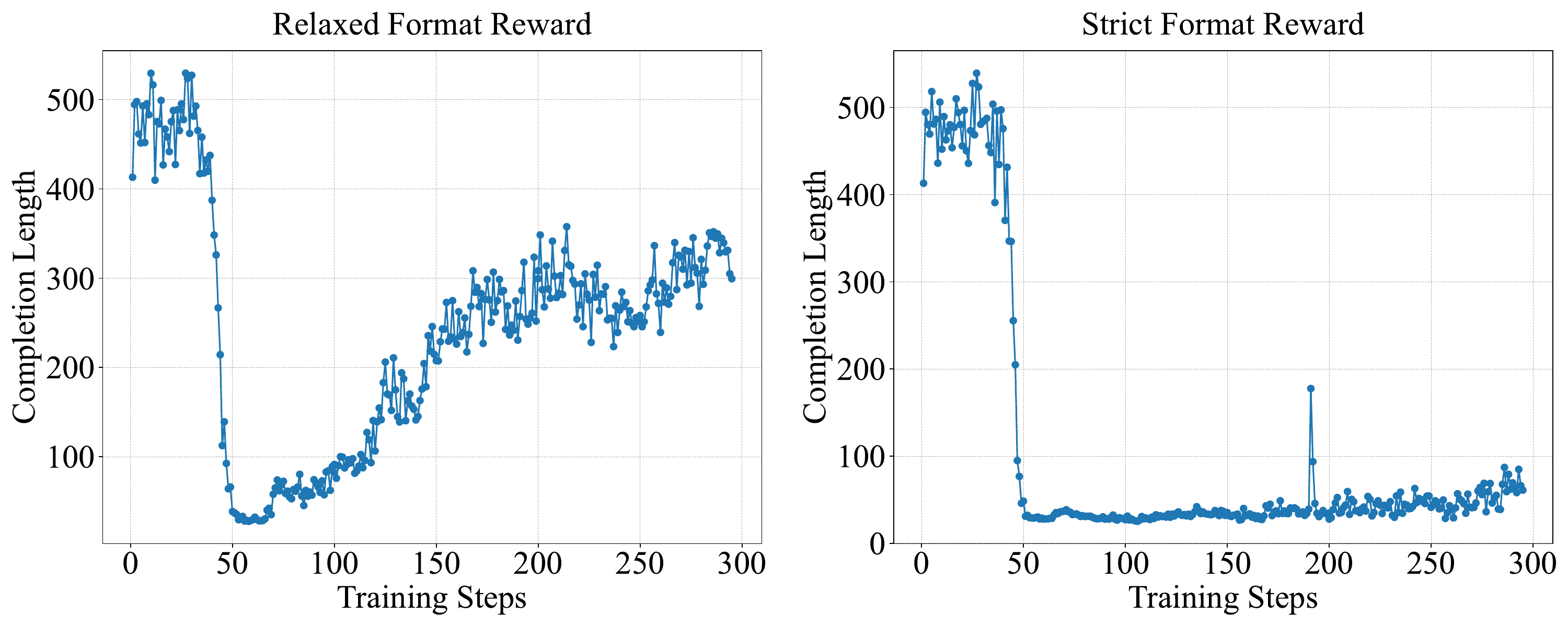}
\caption{Lengths curves of the base model.}
\label{fig:base_compare}
\end{subfigure}
\hfill % 用于在两个子图之间添加间距
\begin{subfigure}[t]{0.48\textwidth}
\centering
\includegraphics[width=6cm]{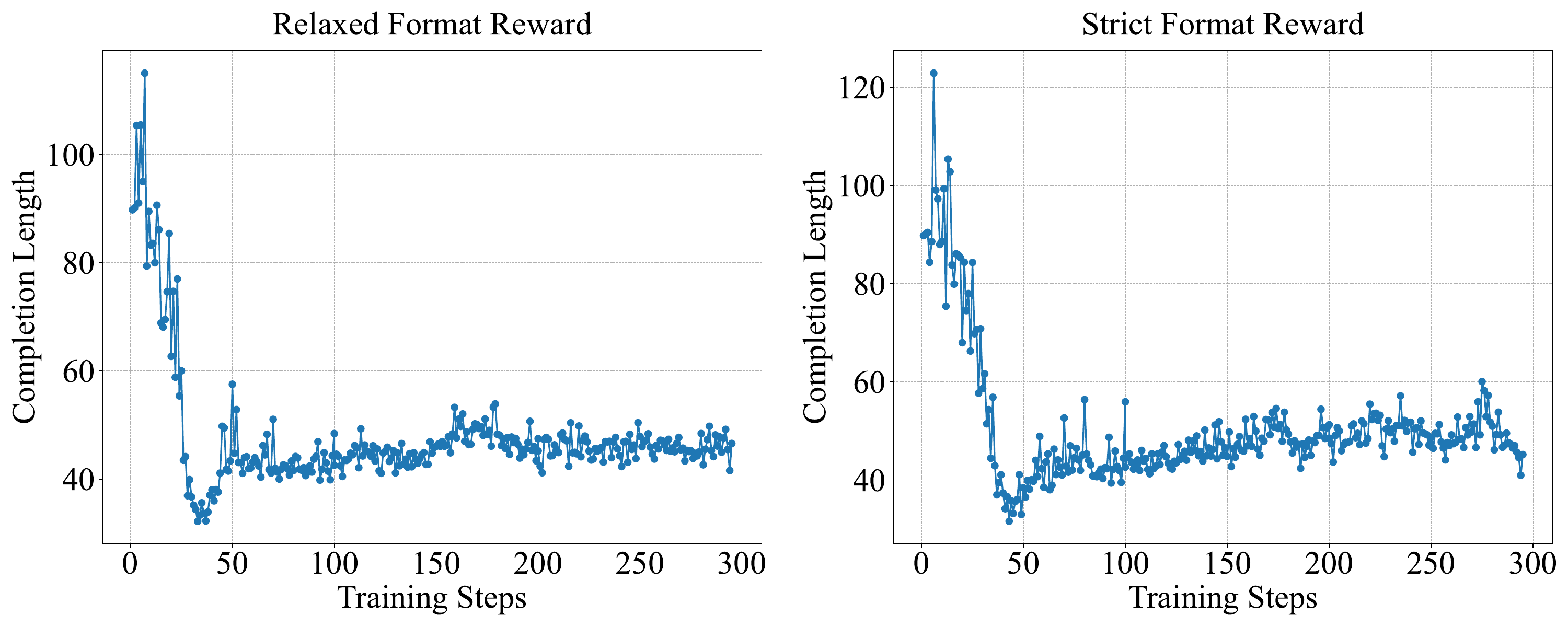}
\caption{Lengths curves of the instruct model.}
\label{fig:instruct_compare}
\end{subfigure}
\caption{Comparison of completion lengths for various models with different format rewards.}
\label{fig:combined_comparison}
\end{figure}

To further investigate the completion lengths of the models and determine whether "aha moments" exist in this task, we reduce the learning rate and increase the training epochs for both models. Additionally, we implement two types of format rewards: 1) A strict format that rigidly restricts the output to the prescribed content, prohibiting any superfluous information; 2) A relaxed format, where the output is deemed correct as long as it encompasses the specified content. As shown in Figure~\ref{fig:base_compare} and~\ref{fig:instruct_compare}, the completion length of the instruct model remains constant under both reward functions. However, the base model displays an initial decrease followed by an increase in completion length under the relaxed format reward. This phenomenon is absent under the stricter format reward. Importantly, the increased length does not contribute valuable information but rather introduces task-irrelevant content. \textbf{This comparison reveals that R1-like reinforcement learning training indeed attempts to increase the length to achieve higher rewards, but true "aha moments" are less likely to emerge in relatively simple intent detection (single-task setting) tasks, as the contextual logic is limited and does not require deep reasoning from the model}.

% \begin{figure}[htbp]
% \centering
% \begin{minipage}[t]{0.48\textwidth}
% \centering
% \includegraphics[width=6.5cm]{figure/base.pdf}
% \caption{Training states of the base model.}
% \label{base_fig}
% \end{minipage}
% \begin{minipage}[t]{0.48\textwidth}
% \centering
% \includegraphics[width=6.5cm]{figure/instruct.pdf}
% \caption{Training states of the instruct model.}
% \label{instruct_fig}
% \end{minipage}
% \end{figure}

\section{Parameter Tuning Tricks}
\begin{figure}[htbp]
\centering
\includegraphics[width=0.5\textwidth]{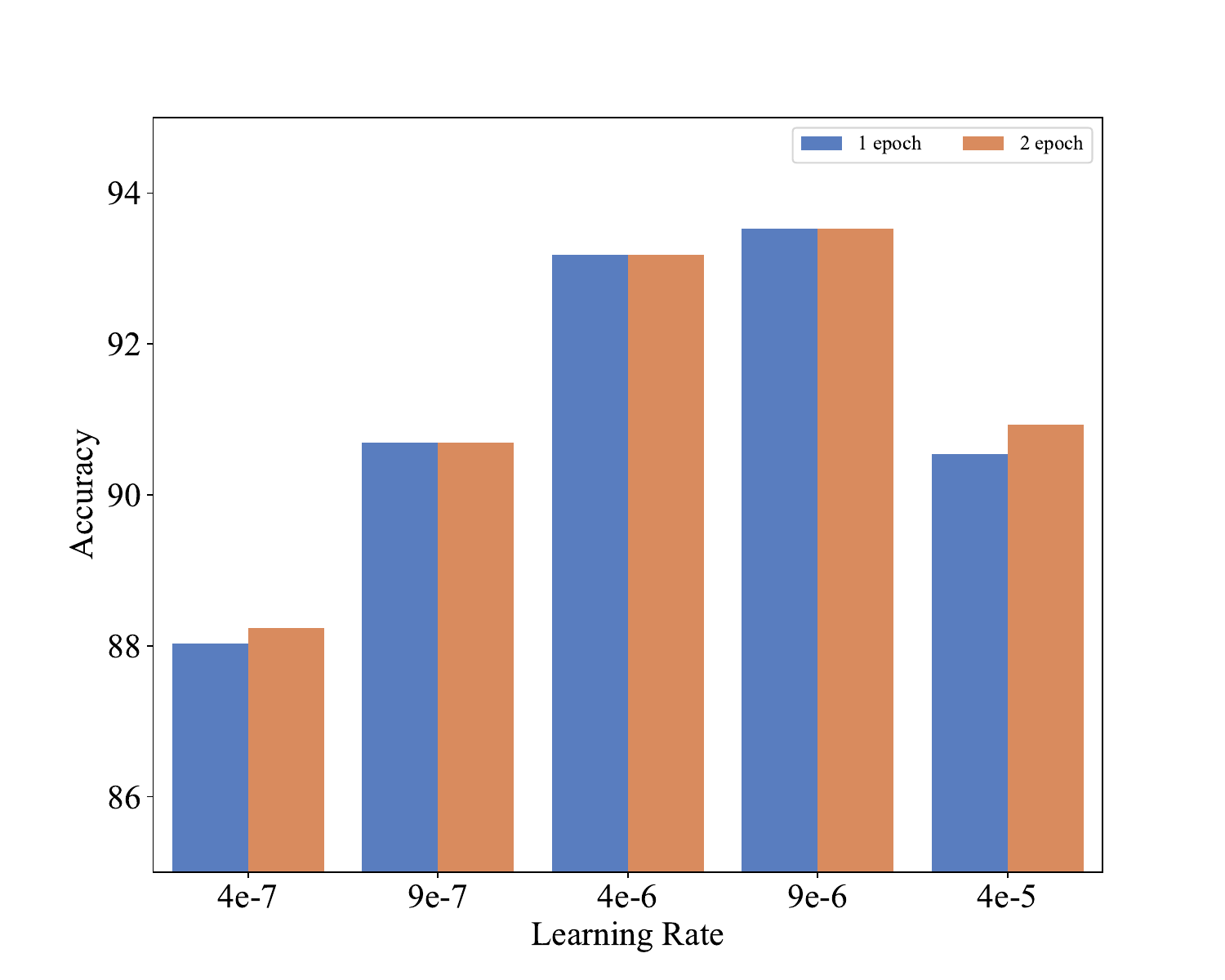} 
\caption{Figure of the accuracy variation with different parameters.} 
\label{params_fig}
\end{figure}

In this section, we will discuss our experimental trials with various parameters in the MultiWOZ dataset. As illustrated in Figure \ref{params_fig}, we conduct experiments with different learning rates. The results indicate that the performance of the model first increases and then decreases as the learning rate increases, achieving optimal performance at a learning rate of $9 \times 10^{-6}$. To investigate whether the low learning rates contributed to the non-convergence of the model, we extend the training for an additional epoch. We observe that increasing the epochs does not improve performance, which demonstrates that one epoch is sufficient for convergence on the intent detection task.

\section{Conclusion}
In this work, to enhance the model's ability to adapt to complex and dynamic scenarios, we apply reinforcement learning to the intent detection task using the GRPO method. We introduce a Reward-based Curriculum Sampling (RCS) method, which leverages the reward function of the GRPO method during the training process to select data of varying difficulty levels. We conduct the curriculum learning approach and sample more challenging data in the second phase. In this way, the model is able to continuously focus on data it does not yet understand, thereby improving its performance and outperforming the SFT method. Furthermore, we empirically demonstrate that the RL-based model exhibits superior generalization capabilities on both in-domain and out-of-domain data. Moreover, we also disclose some interesting findings and share insights regarding parameter tuning encountered during our experimental process.

\section{Next Step}
Moving forward, we intend to channel our research efforts into the following areas:

1) At present, the Reward-based Curriculum Sampling (RCS) we employ is offline. In the future, we plan to transition to an online RCS, which will allow for more efficient selection of superior samples.

2) We aspire to shift our focus from single-intent detection tasks to addressing multi-intent detection tasks, which will significantly improve our capacity to deal with the intricacies of dialogue tasks found in real-world situations.

3) In addition to intent detection tasks, we are set to explore the utilization of reinforcement learning within other facets of Task-Oriented Dialogue (TOD) systems, including but not limited to Dialogue Policy and Response Generation.

4) We are committed to further investigating the deep-seated reasons behind the "aha moment" phenomenon, to augment the task-oriented dialogue model's abilities in self-reflection, self-correction, and self-direction.

\normalem
\bibliographystyle{unsrt}
\bibliography{custom}

\begin{thebibliography}{10}

\bibitem{gupta2024dard}
Aman Gupta, Anirudh Ravichandran, Ziji Zhang, Swair Shah, Anurag Beniwal, and Narayanan Sadagopan.
\newblock Dard: A multi-agent approach for task-oriented dialog systems.
\newblock {\em arXiv preprint arXiv:2411.00427}, 2024.

\bibitem{xu2024rethinking}
Heng-Da Xu, Xian-Ling Mao, Puhai Yang, Fanshu Sun, and He-Yan Huang.
\newblock Rethinking task-oriented dialogue systems: From complex modularity to zero-shot autonomous agent.
\newblock In {\em Proceedings of the 62nd Annual Meeting of the Association for Computational Linguistics (Volume 1: Long Papers)}, pages 2748--2763, 2024.

\bibitem{weld2022survey}
Henry Weld, Xiaoqi Huang, Siqu Long, Josiah Poon, and Soyeon~Caren Han.
\newblock A survey of joint intent detection and slot filling models in natural language understanding.
\newblock {\em ACM Computing Surveys}, 55(8):1--38, 2022.

\bibitem{casanueva2020efficient}
I{\~n}igo Casanueva, Tadas Tem{\v{c}}inas, Daniela Gerz, Matthew Henderson, and Ivan Vuli{\'c}.
\newblock Efficient intent detection with dual sentence encoders.
\newblock {\em arXiv preprint arXiv:2003.04807}, 2020.

\bibitem{du2024anytool}
Yu~Du, Fangyun Wei, and Hongyang Zhang.
\newblock Anytool: Self-reflective, hierarchical agents for large-scale api calls.
\newblock {\em arXiv preprint arXiv:2402.04253}, 2024.

\bibitem{qu2024chatgpt}
Kunyang Qu and Xuande Wu.
\newblock Chatgpt as a call tool in language education: A study of hedonic motivation adoption models in english learning environments.
\newblock {\em Education and Information Technologies}, pages 1--33, 2024.

\bibitem{DBLP:conf/sigir/SiddiqueJXH21}
A.~B. Siddique, Fuad~T. Jamour, Luxun Xu, and Vagelis Hristidis.
\newblock Generalized zero-shot intent detection via commonsense knowledge.
\newblock In Fernando Diaz, Chirag Shah, Torsten Suel, Pablo Castells, Rosie Jones, and Tetsuya Sakai, editors, {\em {SIGIR} '21: The 44th International {ACM} {SIGIR} Conference on Research and Development in Information Retrieval, Virtual Event, Canada, July 11-15, 2021}, pages 1925--1929. {ACM}, 2021.

\bibitem{DBLP:conf/emnlp/ComiCPM23}
Daniele Comi, Dimitrios Christofidellis, Pier~Francesco Piazza, and Matteo Manica.
\newblock Zero-shot-bert-adapters: a zero-shot pipeline for unknown intent detection.
\newblock In Houda Bouamor, Juan Pino, and Kalika Bali, editors, {\em Findings of the Association for Computational Linguistics: {EMNLP} 2023, Singapore, December 6-10, 2023}, pages 650--663. Association for Computational Linguistics, 2023.

\bibitem{parikh2023exploring}
Soham Parikh, Quaizar Vohra, Prashil Tumbade, and Mitul Tiwari.
\newblock Exploring zero and few-shot techniques for intent classification.
\newblock {\em arXiv preprint arXiv:2305.07157}, 2023.

\bibitem{swamy2025roadsleadlikelihoodvalue}
Gokul Swamy, Sanjiban Choudhury, Wen Sun, Zhiwei~Steven Wu, and J.~Andrew Bagnell.
\newblock All roads lead to likelihood: The value of reinforcement learning in fine-tuning, 2025.

\bibitem{guo2025deepseek}
Daya Guo, Dejian Yang, Haowei Zhang, Junxiao Song, Ruoyu Zhang, Runxin Xu, Qihao Zhu, Shirong Ma, Peiyi Wang, Xiao Bi, et~al.
\newblock Deepseek-r1: Incentivizing reasoning capability in llms via reinforcement learning.
\newblock {\em arXiv preprint arXiv:2501.12948}, 2025.

\bibitem{wei2022chain}
Jason Wei, Xuezhi Wang, Dale Schuurmans, Maarten Bosma, Fei Xia, Ed~Chi, Quoc~V Le, Denny Zhou, et~al.
\newblock Chain-of-thought prompting elicits reasoning in large language models.
\newblock {\em Advances in neural information processing systems}, 35:24824--24837, 2022.

\bibitem{shao2024deepseekmath}
Zhihong Shao, Peiyi Wang, Qihao Zhu, Runxin Xu, Junxiao Song, Xiao Bi, Haowei Zhang, Mingchuan Zhang, YK~Li, Y~Wu, et~al.
\newblock Deepseekmath: Pushing the limits of mathematical reasoning in open language models.
\newblock {\em arXiv preprint arXiv:2402.03300}, 2024.

\bibitem{yao2023react}
Shunyu Yao, Jeffrey Zhao, Dian Yu, Nan Du, Izhak Shafran, Karthik Narasimhan, and Yuan Cao.
\newblock React: Synergizing reasoning and acting in language models.
\newblock In {\em 11th International Conference on Learning Representations, ICLR 2023}, 2023.

\bibitem{razin2025makes}
Noam Razin, Zixuan Wang, Hubert Strauss, Stanley Wei, Jason~D Lee, and Sanjeev Arora.
\newblock What makes a reward model a good teacher? an optimization perspective.
\newblock {\em arXiv preprint arXiv:2503.15477}, 2025.

\bibitem{zang2020multiwoz}
Xiaoxue Zang, Abhinav Rastogi, Srinivas Sunkara, Raghav Gupta, Jianguo Zhang, and Jindong Chen.
\newblock Multiwoz 2.2: A dialogue dataset with additional annotation corrections and state tracking baselines.
\newblock {\em arXiv preprint arXiv:2007.12720}, 2020.

\bibitem{hurst2024gpt}
Aaron Hurst, Adam Lerer, Adam~P Goucher, Adam Perelman, Aditya Ramesh, Aidan Clark, AJ~Ostrow, Akila Welihinda, Alan Hayes, Alec Radford, et~al.
\newblock Gpt-4o system card.
\newblock {\em arXiv preprint arXiv:2410.21276}, 2024.

\bibitem{yang2024qwen2}
An~Yang, Baosong Yang, Beichen Zhang, Binyuan Hui, Bo~Zheng, Bowen Yu, Chengyuan Li, Dayiheng Liu, Fei Huang, Haoran Wei, et~al.
\newblock Qwen2. 5 technical report.
\newblock {\em arXiv preprint arXiv:2412.15115}, 2024.

\end{thebibliography}

\end{document}